\documentclass[10pt,twocolumn,letterpaper]{article}

\usepackage{cvpr}
\usepackage{times}
\usepackage{epsfig}
\usepackage{graphicx}
\usepackage{amsmath}
\usepackage{amssymb}
\usepackage{listings}
\usepackage{makecell}
\usepackage{booktabs}
\usepackage{multirow}
\usepackage{color}
\usepackage{comment}
\usepackage{icomma}
\usepackage{enumitem}
\usepackage{amsthm}
\usepackage{authblk}
\setitemize{noitemsep,topsep=0pt,parsep=0pt,partopsep=0pt}
\setenumerate{noitemsep,topsep=0pt,parsep=0pt,partopsep=0pt}
\usepackage[bf,font=footnotesize]{caption}
\usepackage{subfig}
\usepackage{algorithm}
\usepackage[noend]{algorithmic}
\newcommand{\bW}{\mathbf W}

\usepackage[pagebackref=true,breaklinks=true,letterpaper=true,colorlinks,bookmarks=false]{hyperref}

\cvprfinalcopy 

\def\cvprPaperID{185} 

\ifcvprfinal\fi
\begin{document}

\title{Are You Talking to Me? Reasoned Visual Dialog Generation \\through Adversarial Learning}

\author[1]{Qi Wu}
\author[2]{Peng Wang}
\author[1]{Chunhua Shen}
\author[1]{Ian Reid}
\author[1]{Anton van den Hengel}
\affil[1]{Australian Centre for Robotic Vision, The University of Adelaide, Australia}
\affil[2]{Northwestern Polytechnical University, China}

\maketitle

\begin{abstract}
The Visual Dialogue task requires an agent to engage in a conversation about an image with a human.  It represents an extension of the Visual Question Answering task in that the agent needs to answer a question about an image, but it needs to do so in light of the previous dialogue that has taken place.  The key challenge in Visual Dialogue is thus maintaining a consistent, and natural dialogue while continuing to answer questions correctly.  We present a novel approach that combines Reinforcement Learning and Generative Adversarial Networks (GANs) to generate more human-like responses to questions.  The GAN helps overcome the relative paucity of training data, and the tendency of the typical MLE-based approach to generate overly terse answers.  
Critically, the GAN is tightly integrated into the attention mechanism that generates human-interpretable reasons for each answer.  This means that the discriminative model of the GAN has the task of assessing whether a candidate answer is generated by a human or not, given the provided reason.  This is significant because it drives the generative model to produce high quality answers that are well supported by the associated reasoning.
The method also generates the state-of-the-art results on the primary benchmark.
\end{abstract}
\vspace{-10pt}

\section{Introduction}
The combined interpretation of vision and language has enabled the development of a range of applications 
that have made interesting steps towards Artificial Intelligence, including 
Image Captioning \cite{Karpathy2014deepvs,vinyals2014show,wu2015image}, Visual Question Answering (VQA) \cite{antol2015vqa,ren2015image,wu2015ask}, and Referring Expressions \cite{Hu_2016_CVPR,KazemzadehOrdonezMattenBergEMNLP14,yu2016modeling}. 
VQA, for example, requires an agent to answer a previously unseen question about a previously unseen image, and is recognised as being an AI-Complete problem~\cite{antol2015vqa}.
Visual Dialogue~\cite{das2016visual} represents an extension to the VQA problem whereby an agent is required to engage in a dialogue about an image.  This is significant because it demands that the agent is able to answer a series of questions, each of which may be predicated on the previous questions and answers in the dialogue.
Visual Dialogue thus reflects one of the key challenges in AI and Robotics, which is to enable an agent capable of acting upon the world, that we might collaborate with through dialogue.
\begin{figure}[t!]
\centering
  \includegraphics[width=7.8cm]{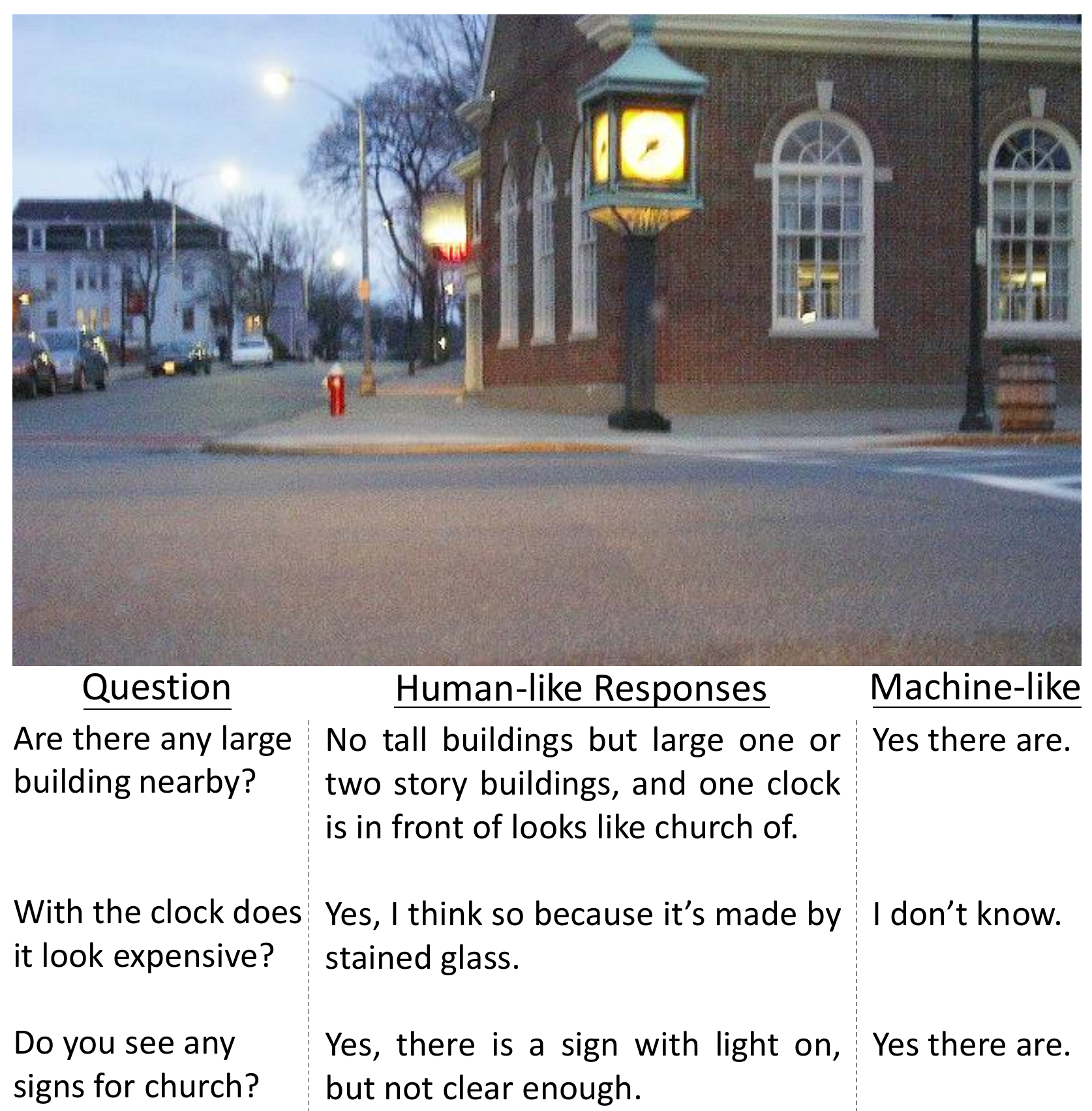}
  \vspace{-3pt}
  \caption{Human-like vs. Machine-like responses in a visual dialog. The human-like responses clearly answer the questions more comprehensively, and help to maintain a meaningful dialogue.}
  \label{img:example}
  \vspace{-10pt}
\end{figure}

Due to the similarity between the VQA and Visual Dialog tasks, VQA methods \cite{lu2016hierarchical,yang2015stacked} have been directly applied to solve the Visual Dialog problem. 
The fact that the Visual Dialog challenge requires an ongoing conversation, however, demands more than just taking into consideration the state of the conversation thus far.
Ideally, the agent should be an engaged participant in the conversation, cooperating towards a larger goal, rather than generating single word answers, even if they are easier to optimise.
Figure~\ref{img:example} provides an example of the distinction between the type of responses a VQA agent might generate and the more involved responses that a human is likely to generate if they are engaged in the conversation.  These more human-like responses are not only longer, they provide reasoning information that might be of use even though it is not specifically asked for.

Previous Visual Dialog systems~\cite{das2016visual} follow a neural translation mechanism that is often used in VQA, by predicting the response given the image and the dialog history using the maximum likelihood estimation (MLE) objective function. However, because this over-simplified training objective only focus on measuring the word-level correctness, the produced responses tend to be generic and repetitive. For example, a simple response of `yes',`no', or `I don't know' can safely answer a large number of questions and lead to a high MLE objective value.
Generating more comprehensive answers, and a deeper engagement of the agent in the dialogue, requires a more engaged training process.

A good dialogue generation model should generate responses indistinguishable from those a human might produce. In this paper, we introduce an adversarial learning strategy, 
motivated by the previous success of adversarial learning in many computer vision  \cite{chen2016infogan,radford2015unsupervised} and sequence generation \cite{dai2017towards, yu2017seqgan} problems. 
We particularly frame the task as
a reinforcement learning problem that we jointly train two sub-modules: a sequence generative model to produce response sentences on the basis of the image content and the dialog history, and a discriminator that leverages previous generator's memories to distinguish between the human-generated dialogues and the machine-generated ones. The generator tends to generate responses that can fool the discriminator into believing that they are human generated, while the output of the discriminative model is used as a reward to the generative model, encouraging it to generate more human-like dialogue.

Although our proposed framework is inspired by generative adversarial networks (GANs) \cite{goodfellow2014generative}, there are several technical contributions that lead to the final success on the visual dialog generation task. First, we propose a sequential co-attention generative model that aims to ensure that attention can be passed effectively across the image, question and dialog history. The co-attended multi-modal features are combined together to generate a response. Secondly, 
and significantly, within the structure we propose the discriminator has access to the attention weights the generator used in generating its response.
Note that the attention weights can be seen as a form of `reason' for the generated response. For example, it indicates which region should be focused on and what dialog pairs are informative when generating the response.
This structure is important as it allows the discriminator to assess the quality of the response, given the reason.  It also allows the discriminator to assess the response in the context of the dialogue thus far.
Finally, as with most sequence generation problems, the quality of the response can only be assessed over the whole sequence.
We follow \cite{yu2017seqgan} to apply Monte Carlo (MC) search to calculate the intermediate rewards.

We evaluate our method on the VisDial dataset \cite{das2016visual} and show that it outperforms the baseline methods by a large margin. We also outperform several state-of-the-art methods. Specifically, our adversarial learned generative model outperforms our strong baseline MLE model by 1.87\% on recall@5, improving over previous best reported results by 2.14\% on recall@5, and 2.50\% recall@10. Qualitative evaluation shows that our generative model generates more informative responses and a human study shows that 49\% of our responses pass the Turing Test. We additionally implement a model under the discriminative setting (a candidate response list is given) and achieve the state-of-the-art performance.

\section{Related work}

\begin{figure*}[t]
\begin{center}
\includegraphics[width=17.5cm]{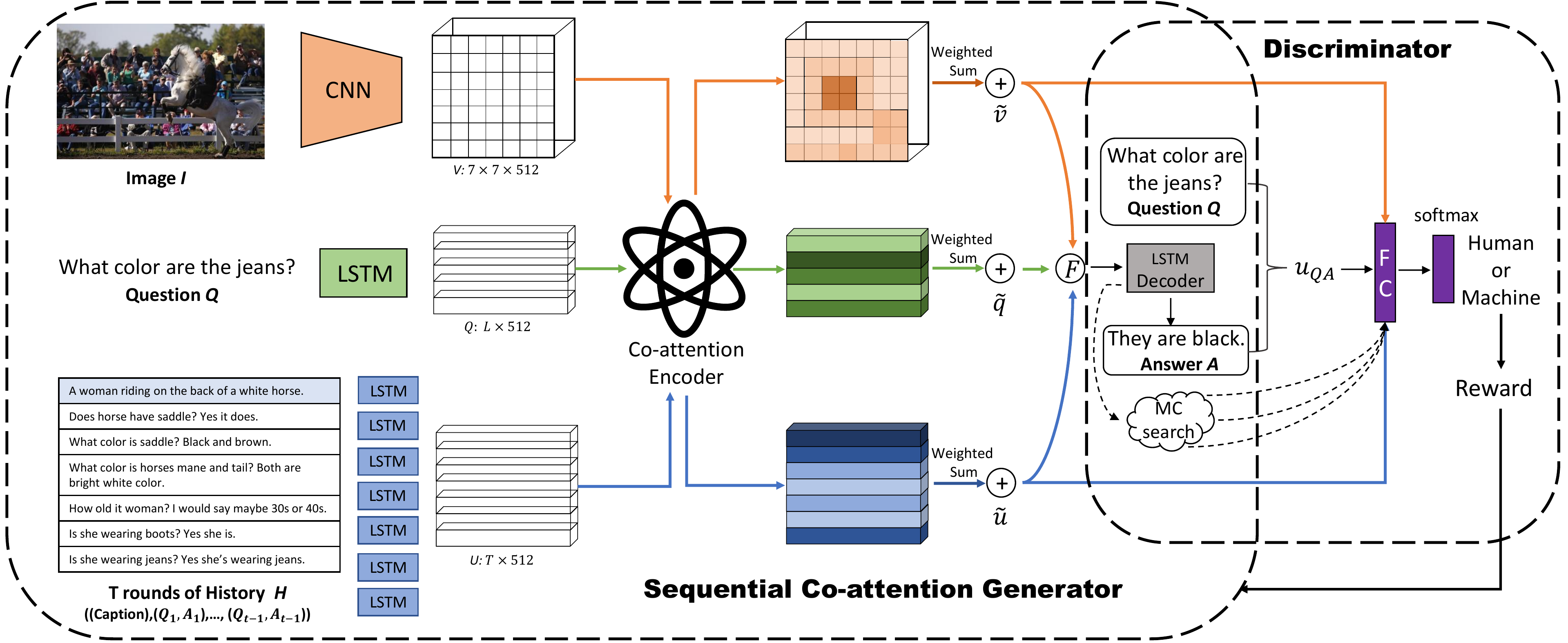}
\end{center}
\vspace{-10pt}
   \caption{The adversarial learning framework of our proposed model. Our model is composed of two components, the first being a sequential co-attention generator that accepts as input \textit{image}, \textit{question} and \textit{dialog history} tuples, and uses the co-attention encoder to jointly reason over them. The second component is a discriminator tasked with labelling whether each answer has been generated by a human or the generative model by considering the attention weights. The output from the discriminator is used as a reward to push the generator to generate responses that are indistinguishable from those a human might generate.}
\vspace{-7pt}
\label{frame}
\end{figure*}

\paragraph{Visual dialog}
is the latest in a succession of vision-and-language problems that began with image captioning \cite{Karpathy2014deepvs,vinyals2014show,wu2015image}, and includes visual question answering \cite{antol2015vqa,ren2015image,wu2015ask}. However, in contrast to these classical vision-and-language tasks that only involve at most a single natural language interaction, visual dialog requires the machine to hold a meaningful dialogue in natural language about visual content. Mostafazadeh \etal \cite{mostafazadeh2017image} propose an Image Grounded Conversation (IGC) dataset and task that requires a model to generate natural-sounding conversations (both questions and responses) about a shared image.  De Vries \etal \cite{de2016guesswhat} propose a ‘GuessWhat’ game style dataset, where one person asks questions about an image to guess which object has been ‘selected’, and the second person answers questions in ‘yes’/‘no’/NA. Das \etal \cite{das2016visual} propose the largest visual dialog dataset, VisDial, by pairing two subjects on Amazon Mechanical Turk to chat about an image. They further formulate the task as a `multi-round' VQA task and evaluate individual responses at each round in a retrieval or multiple-choice setup. 
Recently, Das \etal \cite{das2017learning} propose to use RL to learn the policies of a `Questioner-Bot' and an `Answerer-Bot', based on the goal of selecting the right images that the two agents are talking, from the VisDial dataset. 

Concurrent with our work, Lu \etal \cite{lu2017best} propose a similar generative-discriminative model for Visual Dialog. However, there are two differences. First, their discriminative model requires to receive a list of candidate responses and learns to sort this list from the training dataset, which means the model only can be trained when such information is available.
Second, their discriminator only considers the generated response and the provided list of candidate responses. Instead, we measure whether the generated response is 
valid given the attention weights which reflect both the reasoning of the model, and the history of the dialogue thus far.
As we show in our experiments in Sec. \ref{exp}, this procedure results in our generator producing more suitable responses.

\vspace{-10pt}
\paragraph{Dialog generation in NLP}
Text-only dialog generation \cite{li2016deep,li2017adversarial,ritter2011data,sordoni2015neural,xu2016incorporating} has been studied for many years in the Natural Language Processing (NLP) literature, and has leaded to many applications. Recently, the popular `Xiaoice' produced by Microsoft and the `Its Alive' chatbot created by Facebook have attracted significant public attention. In NLP, dialog generation is typically viewed as a sequence-to-sequence (Seq2Seq) problem, or formulated as a statistical machine translation problem \cite{ritter2011data,sordoni2015neural}. Inspired by the success of the Seq2Seq model \cite{sutskever2014sequence} in the machine translation, \cite{serban2016building,vinyals2015neural} build end-to-end dialog generation models using an encoder-decoder model. 
Reinforcement learning~(RL) has also been applied to train a dialog system. Li \etal \cite{li2016deep} simulate two virtual agents and hand-craft three rewards (informativity, coherence and ease of answering) to train the response generation model. Recently, some works make an effort to integrate the Seq2Seq model and RL. For example, \cite{asghar2016online,su2016continuously} introduce real users by combining RL with neural generation. 

Li~\etal in~\cite{li2017adversarial} were the first to introduce GANs for dialogue generation as an alternative to human evaluation.
They jointly train a generative (Seq2Seq) model to produce response sequences and a discriminator to distinguish between human, and machine-generated responses. Although we also introduce an adversarial learning framework to the visual dialog generation in this work, one of the significant differences is that we need to consider the visual content in both generative and discriminative components of the system, where the previous work \cite{li2017adversarial} only requires textual information. We thus designed a sequential co-attention mechanism for the generator and an attention memory access mechanism for the discriminator so that we can jointly reason over the visual and textual information. Critically, the GAN we proposed here is tightly integrated into the attention mechanism that generates human-interpretable reasons for each answer. It means that the discriminative model of the GAN has the task of assessing whether a candidate answer is generated by a human or not, given the provided reason. This is significant because it drives the generative model to produce high quality answers that are well supported by the associated reasoning. More details about our generator and discriminator can be found in Sections~\ref{generator} and~\ref{discriminator} respectively.

\vspace{-10pt}
\paragraph{Adversarial learning}
Generative adversarial networks \cite{goodfellow2014generative} have enjoyed great successes in a wide range of applications in Computer Vision, \cite{chen2016infogan,radford2015unsupervised,salimans2016improved}, especially in image generation tasks \cite{denton2015deep,zhang2016stackgan}. The learning process is formulated as an adversarial game in which the generative model is trained to generate outputs to fool the discriminator, while the discriminator is trained not to be fooled. These two models can be jointly trained end-to-end. Some recent works have applied the adversarial learning to sequence generation, for example, Yu \etal \cite{yu2017seqgan} backpropagate the error from the discriminator to the sequence generator by using policy gradient reinforcement learning. This model shows outstanding performance on several sequence generation problems, such as speech generation and poem generation. The work is further extended to more tasks such as image captioning \cite{dai2017towards,shetty2017speaking} and dialog  generation \cite{li2017adversarial}. Our work is also inspired by the success of adversarial learning, but we carefully extend it according to our application, \ie the Visual Dialog. Specifically, we redesign the generator and discriminator in order to accept multi-modal information (visual content and dialog history). We also apply an intermediate reward for each generation step in the generator, more details can be found in Sec. \ref{reinforce}.

\section{Adversarial Learning for Visual Dialog Generation}
\label{model}
In this section, we describe our adversarial learning approach to generating natural dialog responses based on an image. There are several ways of defining the visual based dialog generation task \cite{de2016guesswhat,mostafazadeh2017image}. We follow the one in \cite{das2016visual}, in which an image $I$, a `ground truth' dialog history (including an image description $C$) $H = (C, (Q_1, A_1),...,(Q_{t-1}, A_{t-1}))$ (we define each Question-Answer (QA) pair as an utterance $U_t$, and $U_0=C$), and the question $Q$ are given. The visual dialog generation model is required to return a response sentence $\hat{A}=[a_1,a_2,...,a_K]$ to the question, where $K$ is the length (number of words) of the response answer. As in VQA, two types of models may be used to produce the response --- generative and discriminative. In a generative decoder, a word sequence generator (for example, an RNN) is trained to fit the ground truth answer word sequences. For a discriminative decoder, an additional candidate response vocabulary is provided and the problem is re-formulated as a multi-class classification problem. The biggest limitation of the discriminative style decoder is that it only can produce a response if and only if it exists in the fixed vocabulary. 
Our approach is based on a generative model because a fixed vocabulary undermines the general applicability of the model, but also because it offers a better prospect of being extensible to the problem of generating more meaningful dialogue in future. 

In terms of reinforcement learning, our response sentence generation process can be viewed as a sequence of prediction actions that are taken according to a policy defined by a sequential co-attention generative model.  This model is critical as it allows attention (and thus reasoning) to pass across image, question, and dialogue history equally.
A discriminator is trained to label whether a response is human generated or machine generated, conditioned on the image, question and dialog attention memories. Considering here that as we take the dialog and the image as a whole into account, we are actually measuring whether the generated response can be fitted into the visual dialog. The output from this discriminative model is used as a reward to the previous generator, pushing it to generate responses that are more fitting with the dialog history. In order to consider the reward at the local (\ie word and phase) level, we use a Monte Carlo (MC) search strategy and the REINFORCE algorithm \cite{williams1992simple} is used to update the policy gradient. An overview of our model can be found in the Fig.~\ref{frame}. In the following sections, we will introduce each component of our model separately.

\begin{figure}[t]
\begin{center}
\includegraphics[width=8.2cm]{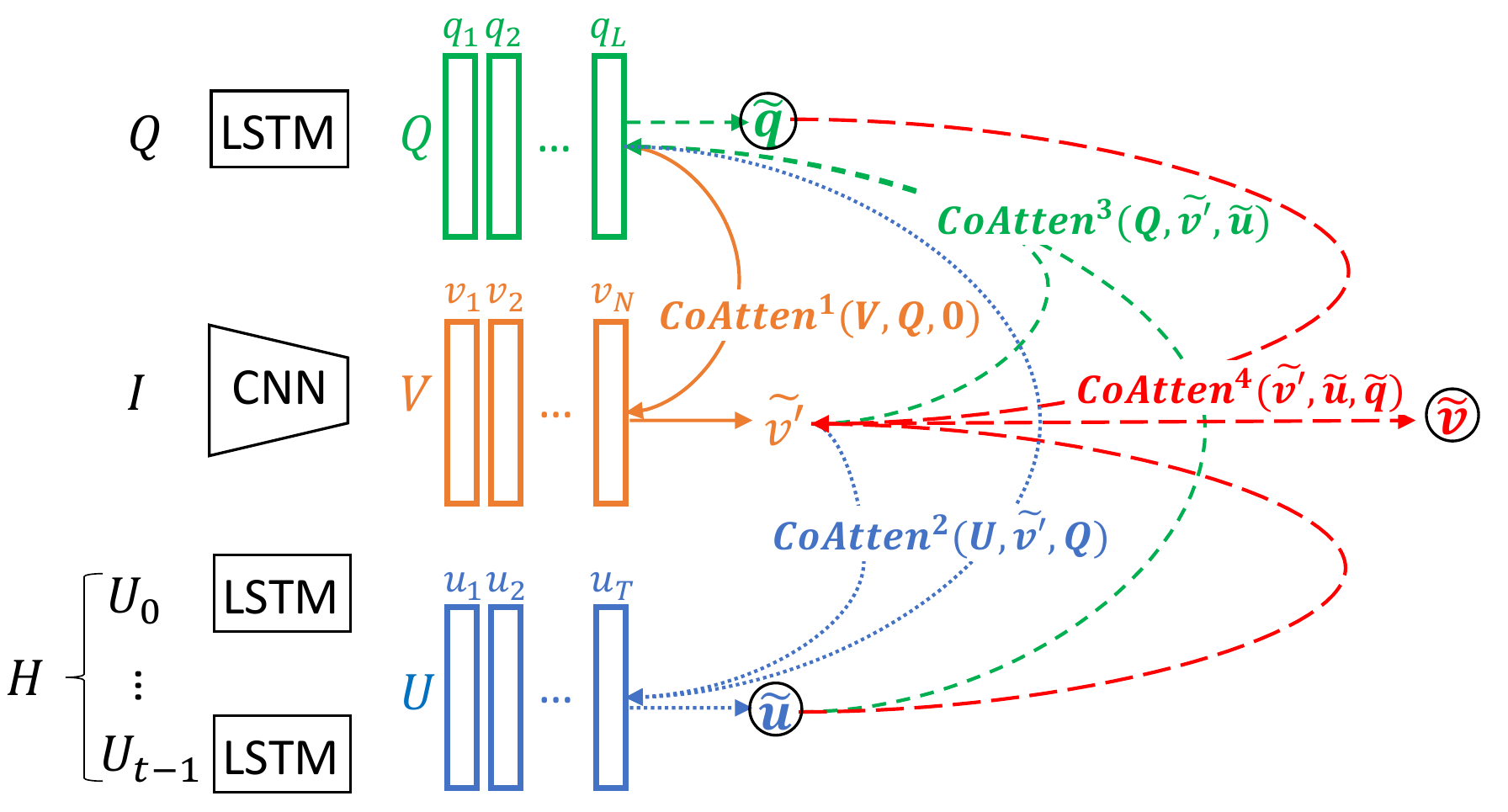}
\vspace{-8pt}
\end{center}
   \caption{The sequential co-attention encoder. Each input feature is co-attend by the other two features in a sequential fashion, using the Eq.1-3. The number on each function indicates the sequential order, and the final attended features $\tilde{u},\tilde{v}$ and $\tilde{q}$ form the output of the encoder.}
\label{coatten}
\vspace{-10pt}
\end{figure}

\subsection{A sequential co-attention generative model}
\label{generator}
We employ the encoder-decoder style generative model which has been widely used in the sequence generation problems. In contrast to text-only dialog generation problem that only needs to consider the dialog history, however, visual dialog generation additionally requires the model to understand visual information. And distinct from VQA that only has one round of questioning, visual dialog has multiple rounds of dialog history that need to be accessed and understood. It suggests that an encoder that can combine multiple information sources is required. A naive way of doing this is to represent the inputs - image, history and question separately and then concatenate them to learn a joint representation. We contend, however, that it is more powerful to let the model selectively focus on regions of the image and segments of the dialog history according to the question. 

Based on this, we propose a sequential co-attention mechanism \cite{wang2016vqa}. Specifically, we first use a pre-trained CNN \cite{simonyan2014very} to extract the spatial image features $V = [v_1, \dots, v_N]$ from the convolutional layer, where $N$ is the number of image regions. The question features is $Q=[q_1, \dots, q_L]$, where $q_l=LSTM(w_l, q_{l-1})$, which is the hidden state of an LSTM at step $l$ given the input word $w_l$ of the question. $L$ is the length of the question. Because the history $H$ is composed by a sequence of utterance, we extract each utterance feature separately to make up the dialog history features, \ie, $U=[u_0, \dots, u_T]$, where $T$ is the number of rounds of the utterance (QA-pairs). And each $u$ is the last hidden state of an LSTM, which accepts the utterance words sequences as the input. 

Given the encoded image, dialog history and question feature $V, U$ and $Q$, we use a co-attention mechanism to generate attention weights for each feature type using the other two as the guidance in a sequential style. Each co-attention operation is denoted as $\tilde{x} = \mathrm{CoAtten}(X, g_1, g_2)$, which can be expressed as follows:
\begin{eqnarray}
\label{eq:seq_attend}
H_i &=& \mathrm{tanh} ( \bW_x x_i \!+\! \bW_{g_1} g_1 \!+\! \bW_{g_2} g_2 ), \\
\alpha_i &=& \mathrm{softmax}(\bW^T H_i), \quad i = 1,\dots,M, \\
\tilde{x}\,\, &=& \textstyle{\sum_{i=1}^{M}} \alpha_i x_i,
\label{catt_que}
\end{eqnarray}
where $X$ is the input feature sequence (\ie, $V$, $U$ or $Q$),
and $g_1$, $g_2 \in \mathbb{R}^d$ represent guidances that are outputs of previous attention modules. Here $d$ is the feature dimension. $\bW_x$, $\bW_{g_1}$, $\bW_{g_2} \in \mathbb{R}^{h \times d}$ and $\bW \in \mathbb{R}^h$ are learnable parameters. Here $h$ denotes the size of hidden layers of the attention module. $M$ is the input sequence length that corresponding to the $N, L$ and $T$ for different feature inputs.

As shown in Fig. \ref{coatten}, in our proposed process, the initial question feature is first used to attend to the image. The weighted image features and the initial question representation are then combined to attend to utterances in the dialog history, to produce the attended dialog history ($\tilde{u}$). The attended dialog history and weighted image region features are then jointly used to guide the question attention ($\tilde{q}$). Finally, we run the image attention ($\tilde{v}$) again, guided by the attended question and dialog history, to complete the circle. All three co-attended features are concatenated together and embedded to the final feature $F$:
\begin{equation}
 F = \mathrm{tanh} (\bW_{eg}[\tilde{v};\tilde{u};\tilde{q}])
\end{equation}
where $[;]$ is a concatenation operator. Finally, this vector representation is fed to an LSTM to compute the probability of generating each token in the target using a softmax function, which forms the response $\hat{A}$. The whole generation process is denoted as $\pi(\hat{A}|V,U,Q)$.

\subsection{A discriminative model with attention memories}
\label{discriminator}
Our discriminative model is a binary classifier that is trained to distinguish whether the input dialog is generated by humans or machines. In order to consider the visual information and the dialog history, we allow the discriminator to access to the attention memories in the generator. Specifically, our discriminator takes $\{\tilde{v}, \tilde{u}, Q, \hat{A}\}$ as the input, where $\tilde{v}, \tilde{u}$ are the attended image and dialog history features produced in the generative model\footnote{we also tested to use the question memory $\tilde{q}$, but we find the discriminator result is not as good as when using the original question input $Q$.}, given the question $Q$. And $\hat{A}$ is the generated response in the generator. The $Q$-$\hat{A}$ pair is further sent to an LSTM to obtain a vector representation $u_{Q\hat{A}}$. All three features are embedded together and sent to a 2-way softmax function, which returns the probability distribution of whether the whole visual dialog is human-natural or not:
\begin{eqnarray}
  \vspace{-8pt}
  O &=& \mathrm{tanh} (\bW_{ed}[\tilde{v};\tilde{u};u_{Q\hat{A}}]) \\
  P &=& \mathrm{softmax} (O) 
  \vspace{-8pt}
\end{eqnarray} 

The probability of the visual dialog being recognised as a human-generated dialog is denoted as $r(\{\tilde{v}, \tilde{u}, Q, \hat{A}\})$.

\subsection{Adversarial REINFORCE with an intermediate reward}
\label{reinforce}
In adversarial learning, we encourage the generator to generate responses that are close to human generated dialogs, or, in our case, we want the generated response can fit into the visual dialog as good as possible. The policy gradient methods are used here to achieve the goal. The probability of the visual dialog being recognised as a human-generated dialog by the discriminator (\ie, $r(\{\tilde{v}, \tilde{u}, Q, \hat{A}\})$) is used as a reward for the generator, which is trained to maximize the expected reward of generated response using the REINFORCE algorithm \cite{williams1992simple}:
\begin{equation}
\vspace{-1pt}
 J(\theta)=\mathbf{E}_{\hat{A}\sim \pi(\hat{A}|V,U,Q)}(r(\{\tilde{v}, \tilde{u}, Q, \hat{A}\})|\theta)
 \label{math:obj}
 \vspace{-1pt}
\end{equation}
Given the input visual information ($V$), question ($Q$) and dialog history utterances ($U$), the generator generates an response answer $\hat{A}$ by sampling from the policy. The attended visual ($\tilde{v}$) and dialog ($\tilde{u}$) memories with the $Q$ and generated answer $\hat{A}$ are concatenated together and fed to the discriminator. We further use the likelihood ratio trick \cite{williams1992simple} to approximate the gradient of Eq. \ref{math:obj}:
\begin{equation}
\begin{split}
 \nabla J(\theta) &\approx \nabla \log \pi(\hat{A}|V,U,Q)\cdot[r(\{\tilde{v}, \tilde{u}, Q, \hat{A}\})-b] \\
 =&\nabla \sum_k \log p(a_k|V,U,Q,a_{1:k-1})\cdot[r(\{\tilde{v}, \tilde{u}, Q, \hat{A}\})-b]
\end{split}
\label{gradient_glo}
\vspace{-5pt}
\end{equation}
where $p$ is the probability of the generated responses words, $a_k$ is the $k$-th word in the response. $b$ denotes the baseline value. Following \cite{li2017adversarial}, we train a critic neural network to estimate the baseline value $b$ by given the current state under the current generation policy $\pi$. The critic network takes the visual content, dialog history and question as input, encodes them to a vector representation with our co-attention model and maps the representation to a scalar. The critic neural network is optimised based on the mean squared loss between the estimated reward and the real reward obtained from the discriminator. The entire model can be trained end-to-end, with the discriminator updating synchronously. We use the human generated dialog history and answers as the positive examples and the machine generated responses as negative examples.

\vspace{-5pt}
\paragraph{Intermediate reward}
An issue in the above vanilla REINFORCE is it only considers a reward value for a finished sequence, and the reward associated with this sequence is used for all actions, \ie, the generation of each token. However, as a sequence generation problem, rewards for intermediate steps are necessary. For example, given a question `\textit{Are they adults or babies?}', the human-generated answer is `\textit{I would say they are adults}', while the machine-generated answer is `\textit{I can't tell}'. The above REINFORCE model will give the same low reward to all the tokens for the machine-generated answer, but a proper reward assignment way is to give the reward separately, \ie, a high reward to the token `\textit{I}' and low rewards for the token `\textit{can't}' and `\textit{tell}'.

Considering that the discriminator is only trained to assign rewards to fully generated sentences, but not intermediate ones, we propose to use the Monte Carlo (MC) search with a roll-out (generator) policy $\pi$ to sample tokens. An N-time MC search can be represented as:
\begin{equation}
 \{\hat{A}^1_{1:K},\dots,\hat{A}^N_{1:K}\}= \text{MC}^\pi(\hat{A}_{1:k};N)
\end{equation}
where $\hat{A}_{1:k}^n=(a_1,\dots,a_k)$ and $\hat{A}_{k+1:K}^n$ are sampled based on the roll-out policy $\pi$ and the current state. We run the roll-out policy starting from the current state till the end of the sequence for $N$ times and the $N$ generated answers share a common prefix $\hat{A}_{1:k}$. These $N$ sequences are fed to the discriminator, the average score 
\begin{equation}
 r_{a_k}=\frac{1}{N}\sum_{n=1}^N r(\{\tilde{v}, \tilde{u}, Q, \hat{A}_{1:K}^n\})
 \label{reward_inter}
\end{equation}
of which is used as a reward for the action of generating the token $a_k$. With this intermediate reward, our gradient is computed as:
\begin{equation}
 \nabla J(\theta)=\nabla \sum_k \log p(a_k|V,U,Q,a_{1:k-1})\cdot[r_{a_k}-b]
 \label{gradient_inter}
\end{equation}
where we can see the intermediate rewards for each generation action are considered.

\begin{algorithm}[t]
\footnotesize
 \caption{Training Visual Dialog Generator with REINFORCE}
 \begin{algorithmic}[1]
  \REQUIRE Pretrained generator $Gen$ and discriminator $Dis$
  \FOR{Each iteration} 
  \STATE $\#$ Train the generator $Gen$
  \FOR{i=1, steps}
  \STATE Sample $(I,H,Q,A)$ from the real data
  \STATE Sample $(\tilde{v},\tilde{u},\hat{A})\sim Gen_{\pi}(\cdot|I,H,Q)$ 
  \STATE Compute Reward $r$ for $(\tilde{v},\tilde{u},Q,\hat{A})$ using $Dis$
  \STATE Evaluate $\nabla J(\theta)$ with Eq. \ref{gradient_glo} or \ref{gradient_inter} depends on whether the intermediate reward (Eq. \ref{reward_inter}) is used
  \STATE Update $Gen$ parameter $\theta$ using $\nabla J(\theta)$
  \STATE Update baseline parameters for $b$
  \STATE Teacher-Forcing: Update $Gen$ on $(I,H,Q,A)$ using MLE
  \ENDFOR
  \STATE $\#$ Train the discriminator $Dis$
  \STATE Sample $(I,H,Q,A)$ from the real data
  \STATE Sample $(\tilde{v},\tilde{u},\hat{A})\sim Gen_{\pi}(\cdot|I,H,Q)$
  \STATE Update $Dis$ using $(\tilde{v},\tilde{u},Q,A)$ as positive examples and $(\tilde{v},\tilde{u},Q,\hat{A})$ as negative examples
  \ENDFOR
 \end{algorithmic}
 \label{alg}
\end{algorithm}

\vspace{-8pt}
\paragraph{Teacher forcing}
Although the reward returned from the discriminator has been used to adjust the generation process, we find it is still important to feed human generated responses to the generator for the model updating. Hence, we apply a teacher forcing \cite{lamb2016professor,li2017adversarial} strategy to update the parameters in the generator. Specifically, at each training iteration, we first update the generator using the reward obtained from the sampled data with the generator policy. Then we sample some data from the real dialog history and use them to update the generator, with a standard maximum likelihood estimation (MLE) objective. The whole training process is reviewed in the Alg. \ref{alg}.

\section{Experiments}
\label{exp}
We evaluate our model on a recently published visual dialog generation dataset, VisDial \cite{das2016visual}. Images in Visdial are all from the MS COCO \cite{lin2014microsoft}, which contain multiple objects in everyday scenes. The dialogs in Visdial are collected by pairing 2 AMT works (a `questioner' and an `answerer') to chat with each other about an image. To make the dialog measurable, the image remains hidden to the questioner and the task of the questioner is to ask questions about this hidden image to ‘imagine the scene better’. The answerer sees the image and his task is to answer questions asked by the questioner. Hence, the conversation is more like multi-rounds of visual based question answering and it only can be ended after 10 rounds. There are 83k dialogs in the COCO training split and 40k in the validation split, for totally 1,232,870 QA pairs, in the Visdial v0.9, which is the latest available version thus far. Following \cite{lin2014microsoft}, we use 80k dialogs for \texttt{train}, 3k for \texttt{val} and 40k as the \texttt{test}.

\subsection{Evaluation Metrics}
Different from the previous language generation tasks that normally use BLEU, MENTOR or ROUGE score for evaluation, we follow \cite{lin2014microsoft} to use a retrieval setting to evaluate the individual responses at each round of a dialog. Specifically, at test time, besides the image, ground truth dialog history and the question, a list of 100 candidates answers are also given. The model is evaluated on retrieval metrics: (1) rank of human response, (2) existence of the human response in top-$k$ ranked responses, \ie, recall@$k$ and (3) mean reciprocal rank (MRR) of the human response. Since we focus on evaluating the generalization ability of our generator, we simply rank the candidates by the generative model's log-likelihood scores.

\begin{figure*}[t]
\centering
\resizebox{\linewidth}{!}{
\begin{tabular}{llp{4.8cm}lp{3cm}lllp{4cm}}
\Xhline{2\arrayrulewidth}
 \hspace{-6pt}\textbf{Image+Caption}&&\hspace{-7pt}\textbf{Question}&&\hspace{-7pt}\textbf{Human Answer}&&\hspace{-7pt}\textbf{CoAtt-G-MLE}&&\hspace{-7pt}\textbf{Ours}\\ \cline{1-1} \cline{3-3} \cline{5-5} \cline{7-7} \cline{9-9}
 \vspace{-8pt}\\
 \scriptsize\begin{tabular}{p{3cm}}\hspace{-6pt}\includegraphics[height=2cm,width=3cm]{./img/523978.jpg}\\\hspace{-10pt}A bathroom with a white bath tub, sink and large window.\end{tabular}&&\scriptsize\hspace{-13pt}\begin{tabular}{p{4.8cm}}What color is the bathroom?\\Are there any people in there?\\Are there towels hanging?\\Is there any soap on the sink?\\What color are the towels?\\What kind of bathtub is it?\\Can you see anything out the bathroom window?\\Are there curtains on the window?\\Is the bathroom light on?\\Is there anything else on the sink?\end{tabular}&&\scriptsize\hspace{-13pt}\begin{tabular}{p{3cm}}The walls are gray\\No\\No folded up\\I do n't think so\\White\\A fancy rectangular\\No\\No\\Yes\\No\end{tabular}&&\scriptsize\hspace{-13pt}\begin{tabular}{l}White\\No\\No\\No soap\\White\\It 's a tub\\No\\No\\Yes\\No\end{tabular}&&\scriptsize\hspace{-13pt}\begin{tabular}{p{4cm}}Most white\\No\\No, on the floor\\I do n't think so\\White\\It 's a shower tub with a shower\\No, just the wall\\No curtains\\Yes\\No\end{tabular}\\ \vspace{-9pt}\\\hline
 \vspace{-8pt}\\ 
 \scriptsize\begin{tabular}{p{3cm}}\hspace{-6pt}\includegraphics[height=2cm,width=3cm]{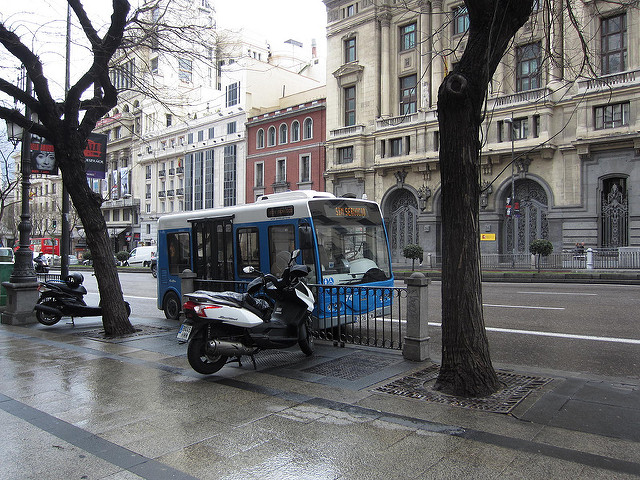}\\\hspace{-10pt}A motorcycle, moped and a bus parked by the street.\end{tabular}&&\scriptsize\hspace{-13pt}\begin{tabular}{p{4.8cm}}What color is the motorcycle?\\Is this on a busy street with shops and people?\\Is it daylight or night time?\\Is the photo in color?\\What color are the other cars?\\Are there any people walking?\\Can you tell what shops businesses they are?\\Do you see any traffic lights?\\Do you think the motorcycle should be parked on the sidewalk?\\Do you see any signs?\end{tabular}&&\scriptsize\hspace{-13pt}\begin{tabular}{p{3cm}}It is black and white\\It looks like it is not\\Daytime\\Yes it is\\I see a white van and a blue\\Not that i can see\\Not really\\No, i do not\\Yes\\One, but only a picture\end{tabular}&&\scriptsize\hspace{-13pt}\begin{tabular}{l}White and blue\\No\\It 's daytime\\Yes\\white and black\\no\\i 'm not sure\\No i do n't\\No\\Yes\end{tabular}&&\scriptsize\hspace{-13pt}\begin{tabular}{p{4cm}}It's black and white\\No it is not\\It is daytime\\Yes\\One is blue and the other is white\\no, there are no people\\I 'm not sure , they are in the background\\No i do n't\\No, it looks like it 's parked\\I see a sign on the side of road\end{tabular}\\ \vspace{-9pt}\\\hline
 \vspace{-8pt}\\ 
 \scriptsize\begin{tabular}{p{3cm}}\hspace{-6pt}\includegraphics[height=2cm,width=3cm]{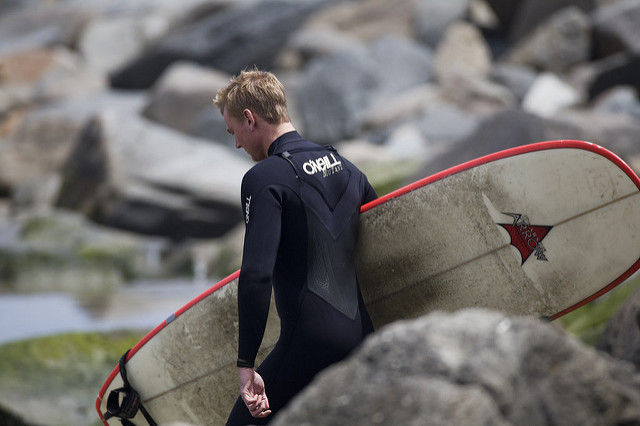}\\\hspace{-10pt}A man in a wet suit carrying a surfboard by some rocks.\end{tabular}&&\scriptsize\hspace{-13pt}\begin{tabular}{p{4.8cm}}Is the photo in color?\\How old does the man appear to be?\\What color wetsuit?\\What color surfboard?\\Do the rocks appear to be smooth or sharp?\\Is he close to the water?\\Does it appear to be a beach or private section?\\What color is the water dark or light blue?\\Does he have any shoes on?\\Does he appear to be wet or dry?\end{tabular}&&\scriptsize\hspace{-13pt}\begin{tabular}{p{3cm}}Yes\\I would estimate late 30s\\Dark blue\\White and red\\I would guess they are smooth\\Moderately close\\Private area\\It is blurry so it appears black\\I ca n't see his feet\\Dry\end{tabular}&&\scriptsize\hspace{-13pt}\begin{tabular}{l}Yes\\20 's\\Black\\White with red\\Smooth\\No \\I ca n't tell\\light blue\\I ca n't see his feet\\Dry\end{tabular}&&\scriptsize\hspace{-13pt}\begin{tabular}{p{4cm}}Yes\\I would say 20 's\\Black\\It 's white with red\\They look smooth\\Yes\\I ca n't tell\\It 's light blue\\I ca n't see his feet\\He looks dry\end{tabular}\\\vspace{-9pt}\\
 \Xhline{2\arrayrulewidth}
\end{tabular}}
   \caption{Qualitative results of our model (\textbf{CoAtt-GAN-w/} $\mathbf R_{inte}$\textbf{-TF}) comparing to human ground-truth answer and our baseline model.}
   \label{qualitative_res}
   \vspace{-4pt}
\end{figure*}

\subsection{Implementation Details}
To pre-process the data, we first lowercase all the texts, convert digits to words, and remove contractions, before tokenizing. The captions, questions and answers are further truncated to ensure that they are no longer than 40, 20 and 20, respectively. We then construct the vocabulary of words that appear at least 5 times in the training split, giving us a vocabulary of 8845 words. The words are represented as one-hot vector and 512-d embeddings for the words are learned. These word embeddings are shared across question, history, decoder LSTMs. All the LSTMs in our model are 1-layered with 512 hidden states. The Adam \cite{kingma2014adam} optimizer is used with the base learning rate of $10^{-3}$, further decreasing to $10^{-5}$. We use 5-time Monte Carlo (MC) search for each token. The co-attention generative model is pre-trained using the ground-truth dialog history for 30 epochs. We also pre-train our discriminator (for 30 epochs), where the positive examples are sampled from the ground-truth dialog, the negative examples are sampled from the dialog generated by our generator. The discriminator is updated after every 20 generator-updating steps.

\subsection{Experiment results}
\paragraph{Baselines and comparative models}
We compare our model with a number of baselines and state-of-the-art models. \textbf{Answer Prior} \cite{das2016visual} is a naive baseline that encodes answer options with an LSTM and scored by a linear classifier, which captures ranking by frequency of answers in the training set. \textbf{NN} \cite{das2016visual} finds the nearest neighbor images and questions for a test question and its related image. The options are then ranked by their mean-similarity to answers to these questions. \textbf{Late Fusion (LF)} \cite{das2016visual} encodes the image, dialog history and question separately and later concatenated together and linearly transformed to a joint representation. \textbf{HRE}~\cite{das2016visual} applies a hierarchical recurrent encoder~\cite{serban2017hierarchical} to encode the dialog history and the \textbf{HREA}~\cite{das2016visual} additionally adds an attention mechanism on the dialogs. \textbf{Memory Network (MN)}~\cite{das2016visual} maintains each previous question and answer as a `fact' in its memory bank and learns to refer to the stored facts and image to answer the question. A concurrent work \cite{lu2017best} proposes a \textbf{HCIAE} (History-Conditioned Image Attentive Encoder) to attend on image and dialog features.

\begin{table}[t]
\centering
\resizebox{\linewidth}{!}{
\begin{tabular}{lccccc}
\Xhline{2\arrayrulewidth}
\textbf{Model}        & \textbf{MRR}    & \textbf{R@1}   & \textbf{R@5}   & \textbf{R@10}  & \textbf{Mean}  \\ \hline
Answer Prior \cite{das2016visual} & 0.3735 & 23.55 & 48.52 & 53.23 & 26.50 \\
NN \cite{das2016visual}           & 0.4274 & 33.13 & 50.83 & 58.69 & 19.62 \\
LF \cite{das2016visual}           & 0.5199 & 41.83 & 61.78 & 67.59 & 17.07 \\
HRE \cite{das2016visual}          & 0.5237 & 42.29 & 62.18 & 67.92 & 17.07 \\
HREA \cite{das2016visual}         & 0.5242 & 42.28 & 62.33 & 68.17 & 16.79 \\
MN \cite{das2016visual}           & 0.5259 & 42.29 & 62.85 & 68.88 & 17.06 \\
HCIAE \cite{lu2017best}        & 0.5386 & 44.06 & 63.55 & 69.24 & 16.01 \\ \hline
CoAtt-G-MLE                              & 0.5411 & 44.32 & 63.82 & 69.75 & 16.47 \\
CoAtt-GAN-w/o $\mathbf R_{inte}$          & 0.5415 & 44.52 & 64.17 & 70.31 & 16.28 \\
CoAtt-GAN-w/ $\mathbf R_{inte}$           & 0.5506 & 45.56 & 65.16 & 71.07 & 15.30 \\
CoAtt-GAN-w/ $\mathbf R_{inte}$-TF        & \textbf{0.5578} & \textbf{46.10} & \textbf{65.69} & \textbf{71.74} & \textbf{14.43} \\ \Xhline{2\arrayrulewidth}
\end{tabular}}
\vspace{-4pt}
\caption{Performance of generative methods on VisDial v0.9. Higher is better for MRR and recall@k, while lower is better for mean rank.}
\vspace{-10pt}
\label{gen-results}
\end{table}

From Table \ref{gen-results}, we can see our final generative model \textbf{CoAtt-GAN-w/ $\mathbf R_{inte}$-TF} performs the best on all the evaluation metrics. Comparing to the previous state-of-the-art model MN \cite{das2016visual}, our model outperforms it by 3.81\% on R@1. We also produce better results than the HCIAE \cite{lu2017best} model, which is the previous best results that without using any discriminative knowledges. Figure \ref{qualitative_res} shows some qualitative results of our model. More results can be found in the supplementary material.

\begin{table}[t]
\centering
\resizebox{\linewidth}{!}{
\begin{tabular}{lccccc}
\Xhline{2\arrayrulewidth}
\textbf{Model}        & \textbf{MRR}    & \textbf{R@1}   & \textbf{R@5}   & \textbf{R@10}  & \textbf{Mean}  \\ \hline
LF \cite{das2016visual}           & 0.5807 & 43.82 & 74.68 & 84.07 & 5.78 \\
HRE \cite{das2016visual}          & 0.5846 & 44.67 & 74.50 & 84.22 & 5.72 \\
HREA \cite{das2016visual}         & 0.5868 & 44.82 & 74.81 & 84.36 & 5.66 \\
MN \cite{das2016visual}           & 0.5965 & 45.55 & 76.22 & 85.37 & 5.46 \\
SAN-QI \cite{yang2015stacked}     & 0.5764 & 43.44 & 74.26 & 83.72 & 5.88 \\
HieCoAtt-QI \cite{lu2016hierarchical} & 0.5788 & 43.51 & 74.49 & 83.96 & 5.84 \\
AMEM \cite{seo2017visual}        & 0.6160 & 47.74 & 78.04 & 86.84 & 4.99 \\
HCIAE-NP-ATT \cite{lu2017best}        & 0.6222 & 48.48 & 78.75 & 87.59 & 4.81 \\ \hline
Ours         & \textbf{0.6398} & \textbf{50.29} & \textbf{80.71} & \textbf{88.81} & \textbf{4.47} \\\Xhline{2\arrayrulewidth}
\end{tabular}}
\vspace{-3pt}
\caption{Performance of discriminative methods on VisDial v0.9. Higher is better for MRR and recall@k, while lower is better for mean rank.}
\label{dis-results}
\vspace{-10pt}
\end{table}

\vspace{-12pt}
\paragraph{Ablation study}
Our model contains several components. In order to verify the contribution of each component, we evaluate several variants of our model.
\begin{itemize}
 \item \textbf{CoAtt-G-MLE} is the generative model that uses our co-attention mechanism shown in Sec. \ref{generator}. This model is trained only with the MLE objective, without any adversarial learning strategies. Hence, it can be used as a baseline model for other variants.
 \item \textbf{CoAtt-GAN-w/o $\mathbf R_{inte}$} is the extension of above CoAtt-G model, with an adversarial learning strategy. The reward from the discriminator is used to guide the generator training, but we only use the global reward to calculate the gradient, as shown in Equ. \ref{gradient_glo}.
 \item \textbf{CoAtt-GAN-w/ $\mathbf R_{inte}$} uses the intermediate reward as shown in the Equ. \ref{reward_inter} and \ref{gradient_inter}.
 \item \textbf{CoAtt-GAN-w/ $\mathbf R_{inte}$-TF} is our final model which adds a `teacher forcing' after the adversarial learning.
\end{itemize}
Our baseline \textbf{CoAtt-G-MLE} model outperforms the previous attention based models (HREA, MN, HCIAE) shows that our co-attention mechanism can effectively encode the complex multi-source information. \textbf{CoAtt-GAN-w/o $\mathbf R_{inte}$} produces slightly better results than our baseline model by using the adversarial learning network, but the improvement is limited. The intermediate reward mechanism contributes the most to the improvement, \ie, our proposed \textbf{CoAtt-GAN-w/~$\mathbf R_{inte}$} model improves over our baseline by average 1\%. The additional Teacher-Forcing model (our final model) brings the further improvement, by average 0.5\%, achieving the best results.

\vspace{-10pt}
\paragraph{Discriminative setting}
We additionally implement a model for the discriminative task on the Visdial dataset \cite{das2016visual}. In this discriminative setting, there is no need to generate a string, instead, a pre-defined answer set is given and the problem is formulated as a classification problem. We modify our model by replacing the response generation LSTM (can be treated as a multi-step classification process) as a single-step classifier. \textbf{HCIAE-NP-ATT} \cite{lu2017best} is the original HCIAE model with a n-pair discriminative loss and a self-attention mechanism. \textbf{AMEM} \cite{seo2017visual} applies a more advanced memory network to model the dependency of current question on previous attention. Additional two VQA models \cite{lu2016hierarchical,yang2015stacked} are used for comparison. Table \ref{dis-results} shows that our model outperforms the previous baseline and state-of-the-art models on all the evaluation metrics.

\subsection{Human study}
Above experiments verify the effectiveness of our proposed model on the Visdial \cite{das2016visual} task. In this section, to check whether our model can generate more human-like dialogs, we conduct a human study.

We randomly sample 1000 results from the test dataset in different length, generated by our final model, our baseline model \textbf{CoAtt-G-MLE}, and the Memory Network (\textbf{MN})\footnote{we use the author provided code and pre-trained model provided on https://github.com/batra-mlp-lab/visdial}~\cite{das2016visual} model. We then ask 3 human subjects to guess whether the last response in the dialog is human-generated or machine-generated and if at least 2 of them agree it is generated by a human, we say it passed the Truing Test. Table \ref{human-results} summarizes the percentage of responses in the dialog that passes the Turing Test (M1), we can see our model outperforms both the baseline model and the MN model. We also apply our discriminator model in Sec. \ref{discriminator} on these 1000 samples and it recognizes that nearly 70\% percent of them as human-generated responses (random guess is 50\%), which suggests that our final generator successfully fool the discriminator in this adversarial learning. We additionally record the percentage of responses that are evaluated as better than or equal to human responses (M2), according to the human subjects' manual evaluation. As shown in Table \ref{human-results}, 45\% of the responses fall into this case.

\begin{table}[t]
\centering
\resizebox{\linewidth}{!}{
\begin{tabular}{>{\centering\arraybackslash}m{5cm} >{\centering\arraybackslash}m{1.2cm} >{\centering\arraybackslash}m{2.5cm} |>{\centering\arraybackslash}m{1.2cm}}
\Xhline{2\arrayrulewidth}
          &MN \cite{das2016visual} &CoAtt-G-MLE&Ours\\ \hline
M1: Percentage of responses that pass the Turing Test            & \large0.39&\large0.46&\large\textbf{0.49} \\\hline
M2: Percentage of responses that are evaluated as better or equal
to human responses.            & \large0.36&\large0.42&\large\textbf{0.45} \\
\Xhline{2\arrayrulewidth}
\end{tabular}}
\vspace{-3pt}
\caption{Human evaluation on 1000 sampled responses on VisDial v0.9}
\label{human-results}
\vspace{-10pt}
\end{table}

\section{Conclusion}

Visual Dialog generation is an interesting topic that requires machine to understand visual content, natural language dialog and have the ability of multi-modal reasoning. More importantly, as a human-computer interaction interface for the further robotics and AI, apart from the correctness, the human-like level of the generated response is a significant index. In this paper, we have proposed an adversarial learning based approach to encourage the generator to generate more human-like dialogs. Technically, by combining a sequential co-attention generative model that can jointly reason the image, dialog history and question, and a discriminator that can dynamically access to the attention memories, with an intermediate reward, our final proposed model achieves the state-of-art on VisDial dataset. A Turing Test fashion study also shows that our model can produce more human-like visual dialog responses.

{\small
\bibliographystyle{ieee}
\bibliography{myref}
}

\end{document}